# Use of Statistical Outlier Detection Method in Adaptive Evolutionary Algorithms

Track: Genetic Algorithms

James M. Whitacre
School of Chemical Engineering
University of New South Wales
Sydney 2052, Australia
+612-9385 5267
z3139475@student.unsw.edu.au

Tuan Q. Pham
School of Chemical Engineering
University of New South Wales
Sydney 2052, Australia
+612-9385 5267
tuan.pham@unsw.edu.au

Ruhul A. Sarker
School of Information Technology and Electrical Engineering, University of New South Wales, ADFA Campus, Canberra 2600, Australia
+612-6268-8051
r.sarker@adfa.edu.au

## ABSTRACT
In this paper, the issue of adapting probabilities for Evolutionary Algorithm (EA) search operators is revisited. A framework is devised for distinguishing between measurements of performance and the interpretation of those measurements for purposes of adaptation. Several examples of measurements and statistical interpretations are provided. Probability value adaptation is tested using an EA with 10 search operators against 10 test problems with results indicating that both the type of measurement and its statistical interpretation play significant roles in EA performance. We also find that selecting operators based on the prevalence of outliers rather than on average performance is able to provide considerable improvements to adaptive methods and soundly outperforms the non-adaptive case.

## Categories and Subject Descriptors
G.1.6 **Optimization**: Stochastic programming.

## General Terms
Algorithms, Measurement, Performance.

## Keywords
Evolutionary Algorithm, Genetic Algorithm, Feedback Adaptation.

## 1. INTRODUCTION
Adaptation of EA parameters is not a new topic with excellent reviews available in [19], [7], [9]. Given the many adaptation methods available and the difficulty of manual parameter tuning, an important task for an EA designer is to select the most effective and user-friendly adaptive method. Listed below are three of the most common strategies for adapting EA parameters.

### 1.1 Common Adaptive Strategies

*1.1.1 Deterministic Methods*
With deterministic methods, parameters are adjusted by an external fixed schedule or by an heuristic based on EA properties during runtime. Although deterministic methods can potentially be very effective, they are also likely to be highly problem-specific and even run-specific and the issue of defining the best deterministic method becomes a complex problem possibly rivaling that of the original optimization problem.

*1.1.2 Self-Adaptive Methods*
With self-adaptive methods, information is encoded in the individual solutions thereby allowing adaptation to occur while the EA is running. However, integrating the tuning/adaptation problem with the original problem's genetic code only acts to combine two largely separable problems making the new search space (solution space + algorithm space) more difficult to search. This is the exact opposite of what should be done for aspects of problem solving that are decomposable.

*1.1.3 Feedback Methods*
Feedback methods use measurements taken on EA performance in order to adapt/control parameters. Unlike self-adaptation where the adaptive mechanism is coupled with the EA, Feedback methods involve an external mechanism that is uncoupled from the search space of the optimization problem. For other discussions on the differences between feedback and self-adaptive methods, see Smith [19].

Since information about the EA is only available on past performance, feedback methods require the assumption that past performance is indicative of current performance. It is also only useful for operators that are executed in parallel and where competition exists between the operators. And, as with the self-adaptive case, there is no assurance that short-term performance gains will result in long-term gains.

Deterministic and self-adaptive methods both increase the complexity of the problem being solved, albeit in different manners. The former creates a secondary problem that must be solved in series with the original problem while the latter creates a secondary problem that can be solved in parallel but still increases the total size of the search space. Feedback methods adapt the parameters online but also separate the parameter search space from the original problem search space making the overall problem less complex. This important advantage makes feedback methods our preferred choice and all future discussion focuses on this approach.

Unfortunately, many of the feedback adaptive methods that have been suggested thus far become difficult to implement when multiple parameters are considered for adaptation. Furthermore, most feedback adaptation heuristics are provided without any justification regarding why they are successful. In light of these shortcomings, this work looks to provide a simple framework for implementing feedback adaptive methods.

## 1.2 Scope of Research

Using the classification proposed by Eiben et al [7], our adaptive method looks to modify the probabilities of search operators (what is changed) using a feedback approach (how the change is made). The changes apply to the entire population (scope of change) and changes occur by monitoring the performance of the operators (evidence for change).

In this work, the performance of a search operator is deduced directly from the competitiveness of the solutions that it generates. For alternative approaches to credit assignment in this context, see [1], [5], [15], [23].

We also assume here that offspring are generated by a single search operator. This is done to both simplify and clarify the process of credit assignment.

Finally, the execution of an operator is defined as resulting in a single offspring. Operators that create multiple offspring will be treated as having a different instance of execution for each offspring.

**Overview**: This work focuses on the types of performance measures available for adaptation and the importance of the statistical interpretation of these measurements. As previously mentioned, the performance of a search operator is directly deduced from the competitiveness of the solutions it creates. Therefore, we shall first consider how this solution competitiveness can be measured in Section 2. Section 3 shows how the performance of an operator can be inferred from competitiveness of the solutions that it has produced. The Experimental Setup is given in Section 4 and Experimental Results from several different adaptive methods are presented in Section 5. Discussion and Conclusions finish off the paper in Sections 6 and 7.

## 2. MEASURING THE PERFORMANCE OF A SOLUTION

The competitiveness of a solution may be calculated in a variety of ways. Listed below are a few examples.

1. **Solution Fitness**
A1: The raw objective function value or fitness of a solution.
A2: The ratio of offspring fitness to parent fitness (similar to fitness improvement).

2. **Raw Selection Pressure**
A3: Binary variable indicating whether offspring solutions survive to the next generation
A4: Binary variable indicating whether parent + offspring solutions survive to the next generation

3. **Solution Age**
A5: The number of selection cycles (ie generations) that a solution survives.

4. **Solution Rank**
A6: The rank of a solution within the population based on objective function value.

Measurements A1 through A6 are just a sample of possible measures and are not meant to be exhaustive. For instance, any of the measures could incorporate parent performance as is done in A2 or placed in some other context (eg niche).

All of the measures listed above are easy to implement but each has potential limitations. A1 and A2 could be sensitive to scaling of the objective function while A3 and A4 have a low resolution which could reduce measurement accuracy. Also, A5 can only be implemented in EA designs where solutions are able to exist over multiple generations. The next section will discuss how these measurements can be used to assess operator performance.

## 3. STATISTICAL INTERPRETATION OF MEASUREMENTS

In fields where experimentation is computer-based, it is rare that a clear distinction is made between measurements and their interpretation. However, it should be recognized that the value of information extracted from measurement data comes both from the measurement apparatus/type and the mathematical sophistication used to manipulate the measurement data. Measures A1 through A6 represent the measurements types that often form the basis of any performance-driven, feedback adaptive method. Any additional manipulation of the performance measurements is typically done without a clear justification .In this work, we encourage the use of operators that have a potential for unexpectedly good offspring assuming that this is more beneficial than encouraging operators that perform well on average.

Our aim is to use statistical tests in order to provide an objective reasoning and straightforward approach to interpreting performance measurements based on the assumptions just stated. Once interpreted, this information can be used for purposes of adaptation. To the best of our knowledge, the only work that is in any way similar to our approach is found in the use of Evolutionary Activity by [2].

In the following discussion, a sample refers to a set of solution performances (termed "measurement data") associated with a particular operator. The statistical measures to be tested include averaging and detection of outliers, as detailed below.

## 3.1 Averaging (I:1)

The average of our sample data is the most common way performance measurements are interpreted. Examples can be seen in [8], [12], [13]. Averaging measurements represents our baseline interpretation method and will be used to compare against other interpretation methods.

## 3.2 Estimating Operator Potential through measurement Outliers (I:3)

In this approach, we look for operators that tend to produce exceptionally good solutions, or outliers. A solution is an outlier when it is NOT expected statistically that the population contains at least one such solution or any better solution. From the distribution of the performance measures, for each solution in the population we can calculate the probability $p_y$ that that solution is an outlier. Summing the $p_y$ of all the solutions produced by an operator gives a measure of the potential of that operator. A quantitative formulation of this definition will now be derived from statistics.

Pooling measurement data from all the operators consistently resulted in a distribution that approximated either a Normal or Log-Normal distribution (based on probability plots). Assuming a Normal distribution is selected, individual measurements $x_i$ are tested against the pooled mean $\mu$ and one-sided $p$ values (eq. 2) are calculated using a $z$ statistic (eq. 1). This $p$ value indicates the probability $p_x$ of observing a measurement of size $x_i$ or greater. Hence, with a simple statistical test, we can determine the extent in which a data point is an outlier. However, we also must account for the fact that the number of outliers will also depend on the size of the sample. This is important because different operators are likely to have different sample sizes.

If there are $n$ measurements in a sample, the number of measurements $\alpha$ of size $x_i$ or greater follows a binomial distribution (eq. 3). The probability $p_y$ of not observing this measurement or a better one after $n$ observations is therefore the probability that $\alpha < 1$ (eq. 4), which can be calculated by the binomial cumulative distribution function. Here the number of observations $n$ refers to the operator sample size.

The final result is $p_y$ which indicates the extent to which a measurement is an outlier that can not be easily accounted for by the stated distribution and the number of points sampled. Summing these $p_y$ values over all solutions produced by an operator provides us with evidence of whether this operator can produce high value outliers beyond that which is expected to occur given the sample size of the operator.[1] The sum of these $p_y$ values for an operator is the definition of I:3. The I:3 statistical interpretation essentially indicates an operator's potential for creating high value (ie outlier) solutions.

$$z_i = \frac{x_i - \mu}{s} \quad (1) \qquad p_x = P(z > z_i) \quad (2)$$

$$\alpha = Bin(n, p_x) \quad (3) \qquad p_y = P(\alpha < 1) \quad (4)$$

Taking a hypothetical sample of data that has been normalized (ie; mean = 0, standard deviation = 1), Figure 1 illustrates how the calculation of $p_y$ will evaluate the measurements for different sample sizes. For each of the sample sizes in Figure 1, the $p_y$ calculation places almost no value on any measurements found below the sample mean. Also notice that for very large measurements ($z > 3$), the $p_y$ calculation approaches a value of 1 meaning we have high confidence that the measurement is an outlier. Finally for values that are large but their classification as outliers is less certain, the sample size from which the measurement is taken will strongly influence the weight placed on the measurement.

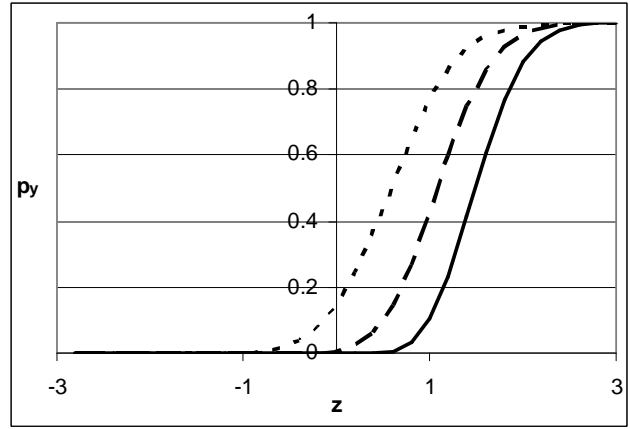

**Figure 1: $p_y$ calculation curves for sample sizes $n$=5 (- - -), $n$=10 (— —), and $n$=20 (——).**

Although statistic I:3 might seem overly complex for those unfamiliar with statistical tests, it is really only a procedure for selectively using measurement outliers for adaptation. Heuristic approaches that also give special treatment to outliers can be found throughout the literature. For instance, adaptive methods such as in [13] only use measurements from offspring if the offspring are better than their parents. Another common approach is to assign a positive evaluation only to the offspring that are better than the median population fitness [15]. Several other heuristics are cited by Barbosa in [1]. A common theme among each of the works cited above is that they give special treatment to a select group of measurements which are to some extent outliers when compared to the other measurements. The difference of the present approach is that we actually quantify the degree to which each solution exceeds the average, and give much more weight to "true" outliers.

### 3.2.1 Applying Statistical Interpretation

When attempting to apply an outlier detection method such as I:3, it is important to recognize that not all measurements are capable of producing outliers. For instance, the measurements which indicate whether or not a solution has survived (A3, A4) would not have the resolution necessary for creating outliers. Also, the scaling of the raw objective function value would make it particularly difficult to observe outliers using common parametric distributions. For these reasons, experimentation using the I:3 interpretation was restricted to adaptive methods using A5 and A6 measurements.

This concludes our discussion of performance measures and the interpretation of those measures. The next section will outline the experimental setup used for testing our adaptive methods.

---

[1] Better statistical approaches might exist for weighting outliers but were not known.

# 4. EXPERIMENTAL SETUP
## 4.1 Search Operators

**Table 1: The 10 search operators used are listed here including name, reference for description, and parameter settings if different from reference.**

| ID | Name | Parameter Settings | Ref |
|---|---|---|---|
| 1 | Wright's Heuristic Crossover | $r = 0.5$[2] | [10] |
| 2 | Simple Crossover | — | [10] |
| 3 | Extended Line Crossover | $\alpha=0.3$[3] | [10] |
| 4 | Uniform Crossover | — | [10] |
| 5 | BLX-$\alpha$ | $\alpha=0.2$ | [10] |
| 6 | Differential Operator | | [20] |
| 7 | Swap | — | — |
| 8 | Raise | $A = 0.01$ | — |
| 9 | Real Number Creep | $A = 0.001$[4] | [17] |
| 10 | Single Point Random Mutation | — | [10] |

Ten search operators were used in all adaptive EA designs as listed in Table 1. Two of the operators were original creations and are described below.

**Swap**: Select the most dissimilar gene between two parents. Transfer all genes from the better fit parent to the offspring except for the previously selected gene which is taken from the less fit parent.

**Raise**: This is similar to Creep except all genes are shifted instead of a single gene. The size of the shift is proportional to the size of each gene's range with $A = 0.01$.

## 4.2 Core EA Design
A real coded EA was used with population size of 30, population solution uniqueness enforced, and binary tournament selection with replacement for both mating (parent) and culling (parent + offspring). Reproduction consisted of the probabilistic use of a single search operator with search operator probabilities normalized to one. Populations were randomly initialized and the stopping criteria was set at a maximum number of generations. All test functions were transformed (if necessary) to be maximization problems with optima at 0. The global optima was assumed to be reached for objective function values > -1E-15.

---

[2] $r$ is set to a static value instead of being a random variable as in the original description

[3] $\alpha$ is set to a static value instead of being a random variable as in the original description

[4] Only a single gene is randomly selected instead of performing operation on all genes.

## 4.3 Test Problems
Experiments were conducted on 10 test problems which are listed in Table 2. All of the test problems used are defined over continuous variable domains with simple bounded constraints on the variables (ie convex search space). The test problems are also characterized as being static with a single objective function. Preliminary experimentation during the design of the adaptive methods occurred using test functions F2, F3, and F4.

**Table 2: Test Problems are listed with identification number, common name, and number of dimensions (variables). More information on each test problem can also be found in the stated reference.**

| ID | Name | Variables | Reference |
|---|---|---|---|
| F1 | Shekel's Foxholes | 2 | [6] |
| F2 | Rastrigin | 20 | [16] |
| F3 | Schwefel | 10 | [16] |
| F4 | Griewank | 10 | [16] |
| F5 | Bohachevsky | 2 | [11] |
| F6 | Watson's | 5 | [11] |
| F7 | Colville's | 4 | [11] |
| F8 | System of linear equations | 10 | [11] |
| F9 | Ackley's | 25 | [11] |
| F10 | Neumaier's #2 | 4 | [14] |

## 4.4 Diversity Control
Single point Mutation (Operator 10) was used for maintaining population diversity. The probability of using operator 10 was set using a deterministic approach proposed by Pham [18] where the probability is exponentially related to the distance $d$ between parents $A$ and $B$. $\delta$ is a parameter that can be tuned and in this work, $\delta = 0.001$ for all test problems.

$$P_{Mut} = P^o_{Mut} + 0.5^{\frac{d}{\delta}} \qquad (5)$$

$$d^2 = \sum_{i=1}^{N_{var}} \left( \frac{x_i^A - x_i^B}{x_{i,max} - x_{i,min}} \right)^2 \qquad (6)$$

In Equation 6, a solution is represented as a vector of search variables $x$ with the $i^{th}$ variable having upper and lower bounds, $x_{i,max}$ and $x_{i,min}$.

## 4.5 Suite of Algorithms Tested
Each adaptive method (A# + I:#) involves the adaptation of all search operator probabilities with probability values updated every 20 generations. Finally, a minimum probability of 0.02 is imposed to ensure a small number of measurements continue to be taken for the worst operators. This is done so that search operators which might be useful later on in the run are not removed completely due to poor initial performance. Probability values are updated so that the new value is 50% from the previous value (Memory) and 50% is from the most current adaptive cycle. All probability values are initialized at equal values unless otherwise stated. We expect little sensitivity to the initial probability settings as was observed in [22]. The use of minimum thresholds [1], [13], [15], [24], adaptation cycles [3], [5], [13], [24], memory [1], and decay parameter [1], [5], [13], [15] are common features of feedback adaptation methods and are seen in the references listed.

Although not thoroughly considered in this work, issues of EA performance degradation from the necessary use of a minimum probability value could be relevant when using a large number of operators [21]. The adaptation cycle length can also be important to performance [4], although in this work, the EA did not appear to be very sensitive to the cycle length (results not shown) as was also largely the case in [22].

**Table 3: Names and adaptive characteristics of EA designs tested. SGA1 and SGA2 are described in more detail elsewhere.**

| Name | Adaptive Measurement | Measurement Interpretation | Diversity Control |
|---|---|---|---|
| SGA1 | N/A | N/A | Yes |
| SGA2 | N/A | N/A | Yes |
| A1-I1 | A1 | I:1 | Yes |
| A2-I1 | A2 | I:1 | Yes |
| A4-I1 | A4 | I:1 | Yes |
| A5-I1 | A5 | I:1 | Yes |
| A5-I3 | A5 | I:3 | Yes |
| A6-I1 | A6 | I:1 | Yes |
| A6-I3 | A6 | I:3 | Yes |

Two EA designs are tested which don't use the feedback adaptive methods discussed. SGA1 uses only two operators: Operator 4 with probability of 0.98 and Operator 10 with probability set by the Diversity Control mechanism ($P^0 = 0.02$). This is included because it is very similar to a standard simple GA. SGA2 uses all 10 operators where operators 1 through 9 are set at probability of 0.1 and operator 10 is set using the Diversity Control mechanism ($P^0 = 0.02$).

## 5. EXPERIMENTAL RESULTS
### 5.1 Assessing EA performance

The performance of a single run of an EA is typically given as the best solution fitness found. Since an EA is a stochastic search process, we must conduct several runs and then extract useful information from the resulting sample of performance data. It is common practice to compare different EA designs based on the sample's average performance and also based on the overall best solution found in a sample. In our results, we decided instead to use a statistical test which measures our confidence that a given EA design is superior to its competitors.

With each EA design being executed 10 times on a test problem, the data set (of EA performances) can be treated as being sampled from a distribution which in turn can be compared with other data sets. Many of the data sets did not fit standard parametric (e.g. $z$ test) distributions as indicated by probability plots and so the Mann-Whitney U Test[5] was used, which doesn't assume a particular distribution.

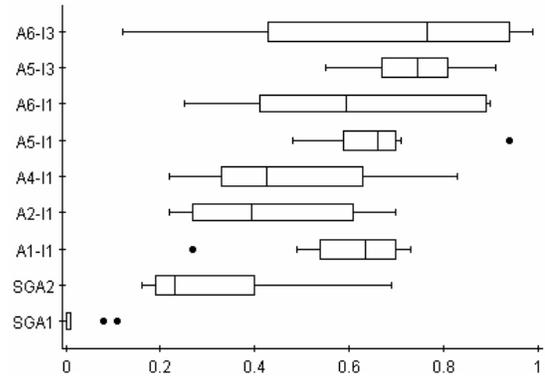

**Figure 2: Boxplot of Mean performance measures for each EA design over all 10 test functions.**

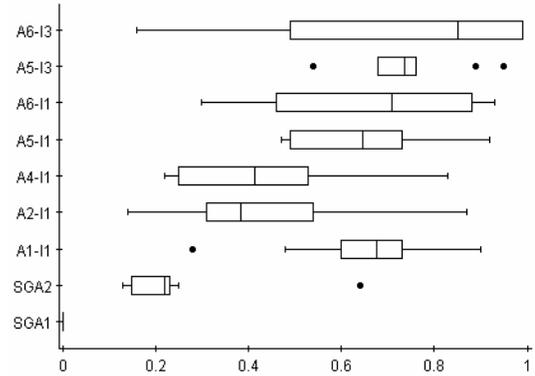

**Figure 3: Boxplot of Final performance measures for each EA design over all 10 test functions.**

Using a one-sided Mann-Whitney U Test, confidence levels (1 - $p$ value) are calculated to indicate whether one sample is greater than another sample. The average confidence level is then used as our measure of EA performance and indicates our expectation that a particular EA design is better than the other competing designs for a particular test problem and stopping criteria.

#### 5.1.1 Stopping Criteria

It is also common for experimental results to state performance for a single stopping criteria. If we want to make general statements about the usefulness of an EA then the selection of a single stopping point will clearly introduce bias into the experimental results as well as the associated conclusions. In an attempt to minimize this bias, stopping points were considered at every 100 generations with a final stopping point at 2000 generations. From the 20 stopping points, three pieces of information have been extracted and are presented in this section.

---

[5] Other non-parametric tests are available and would have also been valid.

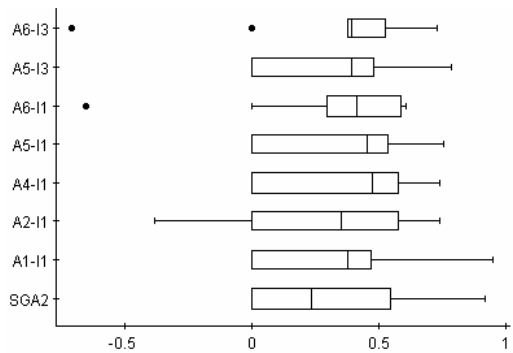

**Figure 4: Boxplot of Correlation Coefficients relating superiority over SGA1 and stopping point.**

First, the average performance over all stopping points (**Mean**) for each EA design is presented in Figure 2. This allows us to see if the EA performed well consistently throughout execution on a specific test problem. Our second measure is simply the performance measure at the final stopping point (**Final**) and is presented in Figure 3. This provides us with some indication of long-term performance. Our last performance measure is a correlation coefficient (**Correlation**). We take the Mann-Whitney statistical test between SGA and each adaptive EA and look for a correlation between this value and the stopping point. This tells us if a particular EA design performs best over short or long time spans in comparison to the simple GA (SGA1).

## 5.2 Analysis of Adaptive Components

**A# Measurement**: An Analysis of Variance (ANOVA) F-test was conducted to determine the extent that performance was affected by the A# measurement type used in adaptation. For this test, only EA designs employing the I:1 interpretation were utilized (A1-I1, A2-I1, A4-I1, A5-I1, A6-I1). The performance measures used as output data in this test were the Mean and Final performance measures. Tests on both performance measures indicate that the measurement type has a significant impact on EA performance (Mean p = 0.039, Final p = 0.010). A summary of the test results is given in Table 4.

**Table 4: ANOVA F-test for significance of A# type**

| Output Measure | Source of Variation | SS | df | MS | F | P-value |
|---|---|---|---|---|---|---|
| Mean | Between Groups | 0.38 | 4 | 0.095 | 2.762 | 0.039 |
| | Within Groups | 1.55 | 45 | 0.035 | | |
| Final | Between Groups | 0.56 | 4 | 0.141 | 3.7413 | 0.010 |
| | Within Groups | 1.69 | 45 | 0.038 | | |

**I# Interpretation**: A paired t-test was conducted to determine whether the use of outliers had a statistically significant positive impact on EA performance. Notice that only A5 and A6 measurements can be used for this test since these were the only two measurements where outliers were observed. For A5, Mean performance results from EA design A5-I1 are paired with A5-I3 for each test function while for A6, results from A6-I1 are paired with A6-I3. Pooling our paired data from A5 and A6 provides us with the results shown in Table 5, which indicate that the selection of statistical outliers has a significant impact on EA performance (p = 0.029). This test was also conducted using Final performance measures, which suggested similar conclusions (p = 0.022).

**Table 5: Paired t-tests for determining statistical significance of improvements from using I:3.**

| Performance Measure | A# | Pairs | n | t | p |
|---|---|---|---|---|---|
| Mean | A5, A6 | A5-I1/A5-I3, A6-I1/A6-I3 | 20 | 2.00 | 0.029 |
| Final | A5, A6 | A5-I1/A5-I3, A6-I1/A6-I3 | 20 | 2.15 | 0.022 |

## 6. DISCUSSION

### 6.1 A# Measurement

The type of measurement (A#) used for indicating the fitness of a solution was found to have a significant impact on both the Mean and Final EA performance measures as indicated by the p-values in Table 4.

Of course the goal of any measurement is to resolve appropriate distinctions between solutions with regards to their reproductive worth. It is not clear what properties attributed to the differences in performance of the A# measurements and we feel that such questions deserve further exploration. It is at least interesting to note that the only two measurements capable of producing detectable outliers were also involved in all of the best EA designs tested.

### 6.2 Interpretation

Among the EA designs tested, it is clear from Figure 2 and Figure 3 that the two designs employing the I:3 outlier detection method (A5-I3 and A6-I3) had superior performance. Particularly impressive was the Mean performance measure of 0.74 for A5-I3. This value means that we have 74% confidence that A5-I3 is the best EA design over all test problems and stopping points considered. This not only suggests strong performance across all the test functions but also strong performance throughout the run for each test function.

A paired t-test for determining the significance of I:3's impact on EA performance (Table 5) suggests that the use of outliers did have a statistically significant effect on both Mean and Final performance measures (p = 0.029 and p = 0.022, resp.).

### 6.3 Adaptation

Probably the easiest trend that can be seen in these results is the immense advantage that adaptation can provide to EA. The only potentially contentious aspect of this conclusion could originate from the bias associated with our core EA design (eg population size) and our somewhat short Final stopping criteria (eg 2000 generations).

For the non-adaptive EA designs, the ten operator design (SGA2) is more effective than the two operator design (SGA1) for every test function considered. Although the focus of this work was not on addressing the utility of implementing large numbers of search operators, the results in Figure 2 and Figure 3 strongly indicate that using a large number of search operators is beneficial, even when not employing an adaptive method for setting the probability values.

## 6.4 Short term vs. Long term Performance

One of the biggest concerns in the EA community with the use of adaptive methods is that for almost all instances, the adaptive methods are designed to favor exploitation.

If an A# measurement is derived based on a solution's fitness, it is rightly anticipated that a highly exploitive search operator will be preferred over an explorative search operator. Although I:3 and other methods [23] have been designed to try to address this problem, we still anticipate that any adaptive method (with A# based on solution fitness) will tend to promote population convergence. Furthermore, it is a generally held belief that encouraging convergence may result in better performance over short time scales but also will likely result in poorer performance over long time scales. Hence, the common conclusion by most EA researchers is that adaptation is only useful for problems where only a small number of function evaluations are possible thereby taking it largely outside of the domain in which EA is successfully applied.

Despite this warning against adaptation, our results indicate the exact opposite is true for all of the adaptive methods tested here. In Figure 4, each of the adaptive methods have their confidence of superiority over SGA1 to be positively correlated to the total number of function evaluations allowed (ie stopping point). The maximum stopping point considered here (2000 generations) is admittedly small relative to some other research in this field however this result does provide an indication that adaptation is not causing poor long term performance.

We readily acknowledge that adaptation based on solution fitness could cause premature convergence for some optimization problems. However we also think that convergence issues would be better addressed by using a structural EA such as Cellular EA as opposed to rejecting adaptation outright as a viable enhancement to EA design.

## 6.5 Limitations to Outlier Detection

It has already been noted that I:3 is only applicable to a measurement that creates detectable outliers.

It also seems reasonable to openly question the assumptions behind the use of I:3. The selection of infrequent high performance measures seems appropriate only to the extent that the computational resources are available to capitalize on them. If the outliers are too infrequent of an occurrence, it may become appropriate to focus attention to other portions of the performance distribution as we approach the end of an optimization run.

Such reasoning is mirrored in successful heuristics such as those used for hybridization of global and local search mechanisms and is also seen in the cooling schedule associated with Simulated Annealing. Automating the transition from exploration to exploitation seems a worthwhile direction of future research.

## 6.6 Practical Implementation Comments

It should be noted that tuning was not attempted for SGA2 while parameter setting for SGA1 was based on experience. There is no doubt that SGA2 would have benefited from parameter tuning. This was not attempted because of the substantial effort required to tune 10 parameters and we assume that this difficulty would make such a design unlikely to be implemented in practice.

Although only probability parameters are considered for adaptation in this work, our approach is not limited to these parameters. Most other EA parameters can also be adapted through their probabilistic implementation or through parallel implementations (e.g. meta-GA). Examples of the former could include finding effective search operator combinations or more generally, the online development of effective search operators from simple building blocks.

Finally, the reader should be reminded that it is possible that some of trends observed here are due to bias from the particular set of operators, test functions, random number seeds, and core EA design selected.

## 7. CONCLUSIONS

The experimental results provide strong evidence that adaptive EA are superior to non-adaptive EA when using a large number of search operators. Although the previous statement was consistently true for all adaptive EAs tested, it is also quite apparent that not all adaptive methods are of equal value.

The selected measure of solution performance was found to be a significant factor in the performance of the adaptive methods. Also significant was the statistical interpretation of performance and particularly the preference of measurement outliers (I:3) was shown to be superior to the averaging of measurements (I:1).